\documentclass[10pt,twocolumn]{article}
\usepackage[T1]{fontenc}
\usepackage[utf8]{inputenc}
\usepackage{lmodern,microtype,amsmath,amssymb,graphicx,booktabs,tabularx,enumitem,xcolor}
\usepackage[letterpaper,margin=0.75in,columnsep=0.25in]{geometry}
\usepackage[hidelinks]{hyperref}
\usepackage{xurl}

\title{Cross-Modal StateKV Injection for Communication in Multi-LLM Systems across Temporal Segments}
\author{%
Xiaoqian Zhang\textsuperscript{1,*}\qquad Zhihao Zhang\textsuperscript{2,3,*}\\[6pt]
\small \textsuperscript{1}Department of Ultrasound \& Faculty of Medical Imaging Technology, Ruijin Hospital,\\
\small College of Health Science and Technology,\\
\small Shanghai Jiao Tong University School of Medicine, Shanghai, China\\[3pt]
\small \textsuperscript{2}School of Mathematics, Shanghai University of Finance and Economics, Shanghai, China\\
\small \textsuperscript{3}School of Computing and Artificial Intelligence,\\
\small Shanghai University of Finance and Economics, Shanghai, China\\[4pt]
\small \texttt{xiaoqian\_shawno@sjtu.edu.cn}\qquad \texttt{2024213167@stu.sufe.edu.cn}\\[3pt]
\small \textsuperscript{*}These authors contributed equally to this work.
}

\date{}
\begin{document}
\maketitle
\begin{abstract}
Video understanding with multiple multimodal language models requires a communication interface that can transfer complementary visual observations. Text-based communication is interpretable, but introduces an intermediate generation stage and can omit information relevant to subsequent reasoning. We study Cross-Modal StateKV Injection, a provider--carrier framework that communicates through key--value (KV) states at the text-token positions that follow the visual input. Provider models observe complementary video frames; because the shared prompt tokens attend to each model's preceding visual tokens, their KV states carry visual-context-conditioned information. The provider's text-span states are projected into the carrier representation space and integrated into the corresponding carrier text-token entries through gated residual updates, leaving the carrier's visual-token entries unchanged. Injection preserves the carrier KV-cache sequence length, while retaining the additional computation required by the providers and the communication module. The reported experiments on MLVU and Video-MME compare carrier-only inference, text-based communication, and StateKV with two provider architectures. Results show modest, task-dependent accuracy gains and lower reported latency than the evaluated text-communication baseline, with some categories remaining unchanged or decreasing. Increasing provider-side frame coverage offers additional gains on selected tasks without extending the carrier cache. These findings motivate further study of KV-state communication of visual context under explicitly controlled observation and computation budgets.
\end{abstract}

\section{Introduction}
Multi-LLM systems enable language models to collaborate through intermediate messages and coordinated reasoning~\cite{r20,r41}. In video understanding, complementary observations may be distributed across temporal segments. A carrier model can reason over its own visual input while using information supplied by another model. The challenge is to communicate that additional evidence without expanding the carrier input indefinitely.

Text-based communication provides a convenient interface~\cite{r22,r38,r46}, but requires a provider to decode an intermediate message and a carrier to consume it. Descriptions may omit visual details needed by a later question. Meanwhile, processing additional frames directly increases the visual token budget. Frame selection and token compression address parts of this trade-off~\cite{r6,r23,r39,r45}, but do not by themselves define a communication interface between independently conditioned models.

We investigate KV states as this interface. Building on Cache-to-Cache communication~\cite{r14}, we study a video setting in which the provider and carrier observe different temporal content. Their visual-token states are segment-specific and are not assumed to be token-aligned. We therefore exchange states at the prompt tokens that follow the video: these tokens are shared across models and, under causal attention, encode each model's preceding visual context. State preparation, feature projection, and gated residual injection are needed before the carrier can use provider information, because shared text does not imply a shared feature space.

Our framework preserves the number of entries in the carrier KV cache: additional observations are incorporated by modifying existing text-token entries rather than appending a provider sequence. This is a carrier-side length constraint, not a claim that the complete multi-model system has constant memory or computation. Provider execution and the injection module remain part of the system cost.

\begin{figure*}[t]
\centering\includegraphics[width=\textwidth]{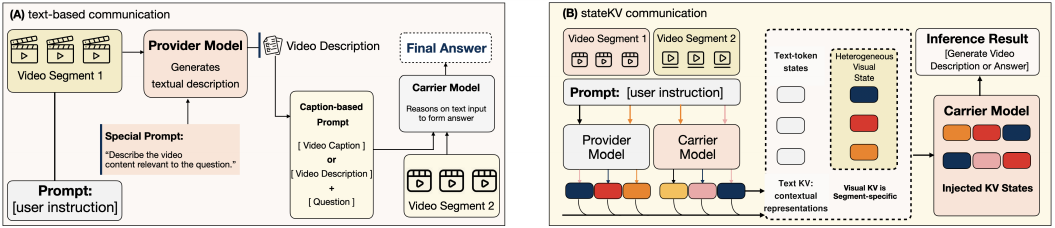}
\caption{Text and KV-state communication. Text communication supplies an intermediate generated description, whereas StateKV transfers visual-context information through KV injection at the shared post-visual text-token positions. A shared textual prompt does not imply identical internal states across models.}
\label{fig:paradigm}
\end{figure*}

The contribution of this study is threefold:
\begin{itemize}[leftmargin=*,nosep]
\item We formulate provider--carrier communication for complementary video observations using the visual-context-conditioned KV states of shared post-visual text tokens.
\item We describe a projected, gated residual injection mechanism that preserves the carrier KV-cache sequence length.
\item We compare the reported accuracy and latency of this interface with carrier-only and text-communication settings on MLVU and Video-MME, including a provider-side frame-budget comparison.
\end{itemize}
The present evidence concerns the evaluated model pairs and frame budgets. It does not establish uniform accuracy gains, lossless semantic transfer, or measured multi-provider parallel speedup.

\section{Related Work}
\subsection{Multi-LLM Communication}
Multi-agent and multi-model systems commonly coordinate through textual messages, roles, or intermediate reasoning traces~\cite{r7,r20,r22,r31,r38,r41,r46}. Such messages can express plans and symbolic conclusions, but their ability to preserve visual evidence depends on the provider prompt and generation budget. Continuous-state communication explores an alternative to that intermediate language interface~\cite{r14,r50}. Our focus is the transfer of complementary video observations through KV representations of visual context.

\subsection{Video Understanding with MLLMs}
Multimodal language models combine visual representations with language reasoning~\cite{r2,r8,r24,r25,r26,r28}. Video input adds temporal coverage and token-budget constraints. Video-MME and MLVU evaluate complementary aspects of video understanding~\cite{r13,r48}, while other benchmarks emphasize multi-shot or long-video reasoning~\cite{r11,r40}. Frame selection, temporal organization, and token reduction offer ways to control visual input costs~\cite{r3,r6,r23,r39,r45}. StateKV instead studies how a carrier can receive information from an additional model while retaining its own cache length.

\subsection{KV Cache and State Communication}
KV reuse and compression seek to reduce inference costs, including attention-based retention, output-perturbation criteria, streaming caches, and reuse across requests~\cite{r9,r12,r30,r43,r44}. These objectives differ from communicating information between separately conditioned models. Cache-to-Cache communication~\cite{r14} is the closest methodological foundation for the present work. Cross-layer KV sharing~\cite{r42} and latent collaboration~\cite{r50} provide related perspectives on internal-state reuse and exchange.

Our intended distinction is the visual communication setting: models observe different video content, so their visual-token sequences are segment-specific. The contribution should therefore be assessed through the concrete choice of exchanging states at shared post-visual text tokens and the associated preparation procedure, rather than treating KV communication itself as a new primitive. Shared natural-language input is not sufficient evidence for a shared cross-model latent coordinate system.

\section{Methodology}
\subsection{Problem Setting}
Let $X$ be a textual prompt and let $\mathcal F_c$ and $\mathcal F_p^{(m)}$ denote the frames observed by the carrier and provider $m$, respectively. We allow $M$ providers in the notation, although the reported tables evaluate one provider at a time. Frame sets represent the visual observations assigned to the two roles.

Each model receives its frames followed by the same textual prompt. At carrier layer $\ell$, the prefill cache is $(K_c^\ell,V_c^\ell)\in\mathbb R^{N_c\times H_c\times d_c}$. Provider $m$ supplies $(K_p^{m,j},V_p^{m,j})\in\mathbb R^{N_p^m\times H_p^m\times d_p^m}$ at layer $j$. Here $N$ is the token count, $H$ the number of KV heads, and $d$ the per-head dimension. Layer counts are denoted $D_c,D_p^m$ to avoid confusing depth with sequence length.

\subsection{Text-Span State Preparation}
The communication interface is the KV state at the \emph{text-token positions that follow the visual input}. Visual-token KV entries are segment-specific: providers and carrier observe different frames, and their visual sequences need not correspond position by position. In contrast, the prompt tokens placed after the video are the same for all models. Under causal attention, their key and value states are computed after attending to the preceding visual tokens, and therefore carry a visual-context-conditioned representation of what each model has observed. We use these positions as the exchange interface; the carrier's own visual-token entries are left unchanged.

Let $e_c$ and $e_p^m$ be the positions of the final visual-end marker in the carrier and provider inputs, as determined by the actual input construction. The transferred span is
\begin{equation}
\begin{aligned}
I_c&=\{e_c+1,\ldots,e_c+N_t\},\\
I_p^m&=\{e_p^m+o_m+1,\ldots,e_p^m+o_m+N_t\},
\end{aligned}
\end{equation}
where $N_t$ is the number of post-visual prompt tokens and $o_m$ is a model-specific offset for separator tokens that some chat templates insert after the visual block ($o_m=0$ for Qwen3-VL). We define
\begin{align}
(K_{c,t}^\ell,V_{c,t}^\ell)&=\operatorname{Select}_{I_c}(K_c^\ell,V_c^\ell),\\
(K_{p,t}^{m,j},V_{p,t}^{m,j})&=\operatorname{Select}_{I_p^m}(K_p^{m,j},V_p^{m,j}).
\end{align}
Because the prompt text is shared and the evaluated models use compatible tokenizers, the two spans contain the same token sequence and are aligned one-to-one; no truncation, interpolation or learned sequence alignment is applied. Shared text does not imply identical KV states: the provider and carrier states differ in feature space and in the visual context they condition on, which is why a learned projection is required. Keys are exchanged in their cached form, after rotary position encoding, without explicit re-rotation.

\begin{figure*}[t]
\centering\includegraphics[width=\textwidth]{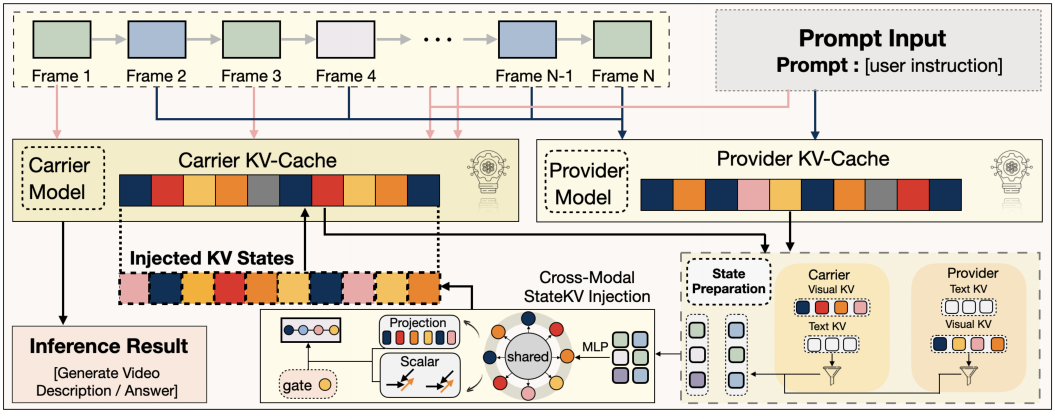}
\caption{StateKV architecture. State preparation separates visual and text KV entries and selects the post-visual text-token span from the provider and carrier caches. A projector with a shared MLP, a projection branch, a per-head scalar branch and a learned gate produces a residual update that is written back to the carrier's text-token positions. Carrier visual-token entries are not modified. The carrier performs the final autoregressive decoding.}
\label{fig:architecture}
\end{figure*}

\subsection{Layer Alignment and Residual Injection}
Using one-based indices, the top-aligned layer mapping is
\begin{equation}
j=D_p^m-D_c+\ell,\qquad \ell=1,\ldots,D_c,
\end{equation}
for $D_p^m\geq D_c$. This aligns layers by relative depth; semantic equivalence is not assumed. The reported 36-to-28-layer configuration maps carrier layers 1--28 to provider layers 9--36 (zero-based 8--35).

For each carrier layer, a projector receives the aligned provider and carrier text-span states. Head and feature axes are flattened per token, giving $\bar K_{p,t},\bar V_{p,t}\in\mathbb R^{N_t\times H_pd_p}$ and $\bar K_{c,t},\bar V_{c,t}\in\mathbb R^{N_t\times H_cd_c}$. We describe the key path; the value path is identical with separate parameters. A shared representation is formed from the token-wise concatenation:
\begin{equation}
h^\ell=\operatorname{MLP}_1^\ell\!\left(W_{\mathrm{in}}^\ell[\bar K_{p,t};\bar K_{c,t}]\right).
\end{equation}
A projection branch predicts an update in the carrier key space,
\begin{equation}
P_K^\ell=W_{\mathrm{out}}^\ell\,\operatorname{MLP}_2^\ell(h^\ell)\in\mathbb R^{N_t\times H_cd_c},
\end{equation}
which is reshaped into carrier KV heads. A scalar branch predicts a per-token, per-head weight
\begin{equation}
s_K^\ell=\sigma\!\left(W_s^\ell\,\operatorname{MLP}_3^\ell(h^\ell)\right)\in(0,1)^{N_t\times H_c},
\end{equation}
broadcast over the head dimension. Each layer also has a learned scalar gate $g_K^\ell=\sigma(\gamma_K^\ell)$. During training, Gumbel noise is added to $\gamma_K^\ell$ and the gate temperature is annealed; at inference the deterministic sigmoid is used. The update is an additive residual with a fixed scale $\alpha$:
\begin{equation}
\widehat K_{c,t}^\ell=K_{c,t}^\ell+\alpha\,g_K^\ell\,s_K^\ell\odot P_K^\ell,\qquad \alpha=0.05,
\end{equation}
and analogously for $\widehat V_{c,t}^\ell$. The updated entries are written back to positions $I_c$; all other carrier cache entries, including visual tokens, are unchanged and the total cache length remains $N_c$. The carrier then processes the remaining prompt token(s) and decodes from this modified cache. The provider backbone and carrier backbone are frozen; only the projectors are trained, using the carrier's next-token cross-entropy on the answer.

\subsection{Carrier-Side Budget}
For a fixed carrier input and cache dtype, rewriting existing KV entries preserves the carrier cache shape. Total system memory also includes the provider weights, provider cache, projector, and intermediate buffers. Likewise, provider prefill and state transfer contribute to inference latency. We therefore distinguish carrier cache length from total computation and memory throughout the evaluation.

\section{Experiments}
\subsection{Evaluation Setup}
We evaluate on MLVU~\cite{r48} and Video-MME~\cite{r13}, using Qwen3-VL-2B-Instruct as the carrier and Qwen3-VL-4B-Instruct or InternVL3.5-4B as the provider. The Qwen models use scaled dot-product attention (SDPA), while InternVL uses eager attention. MLVU is evaluated on its development set, and Video-MME is evaluated without subtitles.

Three settings are compared: carrier-only inference; text communication, in which the provider generates an intermediate description; and StateKV communication. The main tables use eight frames for each role. A separate comparison increases provider input from eight to sixteen frames while keeping the carrier at eight frames.

We report accuracy by task and video-duration group. Timing is listed separately from accuracy to distinguish prediction quality from inference cost.

\subsection{MLVU Results}
Tables~\ref{tab:mlvu-qwen} and~\ref{tab:mlvu-intern} show task-dependent effects. With the Qwen provider, StateKV improves counting, ego reasoning, needle retrieval, ordering and plot QA over carrier-only inference, while anomaly recognition and topic reasoning decrease. With the InternVL provider, improvements occur on five of the seven categories; needle retrieval and anomaly recognition decrease.

The unweighted means computed from the displayed seven category scores are 48.97\% for the carrier, 49.65\% for Qwen-provider StateKV, and 49.52\% for InternVL-provider StateKV. These are unweighted summaries of rounded task-level entries. The observed gains are modest and do not establish uniform improvement or statistical significance.

\subsection{Video-MME Results}
Tables~\ref{tab:mme-qwen} and~\ref{tab:mme-intern} separate short, medium and long videos. Qwen-provider StateKV increases long-video accuracy from 41.56\% to 42.22\%, but decreases short-video accuracy from 55.67\% to 55.44\% and medium-video accuracy from 43.11\% to 42.89\%. InternVL-provider StateKV leaves short-video accuracy unchanged, decreases medium-video accuracy, and improves long-video accuracy to 42.44\%.

The corresponding simple means of the three displayed duration scores are 46.78\%, 46.85\% and 47.00\%, respectively. These unweighted duration-group means are distinguished from example-weighted overall accuracy.

\subsection{Communication Cost and Frame Coverage}
\begin{figure*}[tp]\centering
\includegraphics[width=\textwidth]{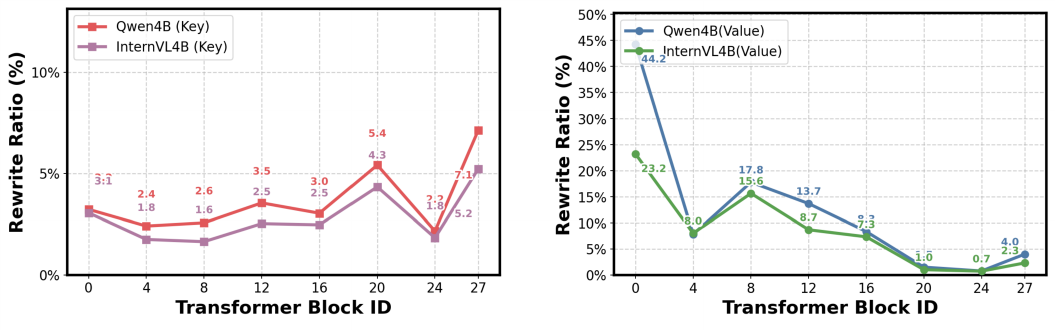}
\caption{Layer-wise rewrite ratios for keys (left) and values (right). The carrier is Qwen3-VL-2B; providers are Qwen3-VL-4B and InternVL3.5-4B.}\label{fig:rewrite}
\end{figure*}

Across the reported tables, StateKV has lower measured time than text communication and higher measured time than carrier-only inference. This comparison concerns the specified baselines; it does not imply that KV injection removes provider computation. Figure~\ref{fig:rewrite} illustrates layer-wise rewrite ratios for keys and values, measured as the relative magnitude of the update on the injected text-token span. The layer-dependent patterns describe the magnitude of cache modification; they do not directly measure the causal contribution of each layer to prediction accuracy.

Table~\ref{tab:frames} compares eight and sixteen provider frames with eight carrier frames. Additional provider frames improve some categories and reduce others. The result motivates testing complementary temporal coverage under controlled observation budgets. Because the carrier-only baseline receives fewer total model observations, these experiments alone cannot isolate the communication mechanism from the benefit of additional observations. A matched-observation control and a matched-compute comparison would make that attribution clearer.

\begin{figure*}[tp]\centering
\includegraphics[width=\textwidth]{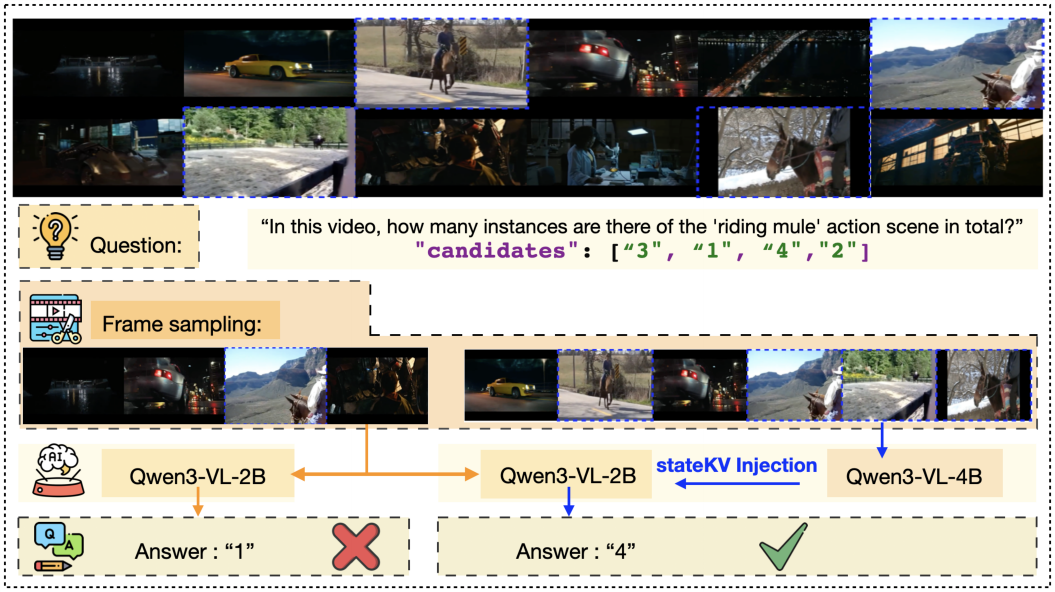}
\caption{Qualitative counting example. Carrier-only inference predicts one riding-mule instance, while the provider--carrier setting predicts four after observing additional frames. The additional observations are incorporated through KV injection at the carrier's post-visual text-token positions while retaining the carrier cache length. This example illustrates the information pathway rather than establishing an aggregate accuracy gain.}\label{fig:temporal}
\end{figure*}
\begin{table*}[tp]\centering\small
\caption{MLVU with the Qwen provider.}\label{tab:mlvu-qwen}
\begin{tabular}{lrrrrrr}\toprule
 & \multicolumn{3}{c}{Accuracy (\%)} & \multicolumn{3}{c}{Reported time (s)}\\
Task & Carrier & Text & StateKV & Carrier & Text & StateKV\\\midrule
Count & 17.48 & 14.56 & 20.87 & 0.19 & 3.22 & 0.43\\
Ego & 46.38 & 43.18 & 46.88 & 1.17 & 5.11 & 1.9\\
Needle & 49.29 & 40.84 & 49.39 & 0.21 & 3.26 & 0.46\\
Order & 35.90 & 33.97 & 38.22 & 0.4 & 3.43 & 0.71\\
Plot QA & 49.92 & 45.64 & 50.72 & 0.64 & 4.19 & 1.23\\
Anomaly recognition & 64.00 & 57.50 & 62.00 & 0.13 & 3.17 & 0.33\\
Topic reasoning & 79.85 & 76.42 & 79.47 & 0.68 & 4.121 & 1.21\\
\bottomrule\end{tabular}\end{table*}
\begin{table*}[tp]\centering\small
\caption{MLVU with the InternVL provider.}\label{tab:mlvu-intern}
\begin{tabular}{lrrrrrr}\toprule
 & \multicolumn{3}{c}{Accuracy (\%)} & \multicolumn{3}{c}{Reported time (s)}\\
Task & Carrier & Text & StateKV & Carrier & Text & StateKV\\\midrule
Count & 17.48 & 16.01 & 19.90 & 0.19 & 12.83 & 0.86\\
Ego & 46.38 & 42.26 & 48.01 & 1.17 & 15.2 & 1.32\\
Needle & 49.29 & 40.56 & 48.73 & 0.21 & 13.97 & 0.91\\
Order & 35.90 & 32.05 & 36.68 & 0.4 & 14.17 & 1.14\\
Plot QA & 49.92 & 45.40 & 50.72 & 0.64 & 14.53 & 1.08\\
Anomaly recognition & 64.00 & 59.00 & 62.00 & 0.13 & 11.9 & 0.79\\
Topic reasoning & 79.85 & 75.29 & 80.60 & 0.68 & 13.94 & 1.06\\
\bottomrule\end{tabular}\end{table*}
\begin{table*}[tp]\centering\small
\caption{Video-MME with the Qwen provider.}\label{tab:mme-qwen}
\begin{tabular}{lrrrrrr}\toprule
 & \multicolumn{3}{c}{Accuracy (\%)} & \multicolumn{3}{c}{Reported time (s)}\\
Task & Carrier & Text & StateKV & Carrier & Text & StateKV\\\midrule
Short & 55.67 & 55.89 & 55.44 & 0.23 & 3.61 & 0.59\\
Medium & 43.11 & 42.56 & 42.89 & 0.22 & 3.72 & 0.59\\
Long & 41.56 & 39.00 & 42.22 & 0.23 & 3.7 & 0.6\\
\bottomrule\end{tabular}\end{table*}
\begin{table*}[tp]\centering\small
\caption{Video-MME with the InternVL provider.}\label{tab:mme-intern}
\begin{tabular}{lrrrrrr}\toprule
 & \multicolumn{3}{c}{Accuracy (\%)} & \multicolumn{3}{c}{Reported time (s)}\\
Task & Carrier & Text & StateKV & Carrier & Text & StateKV\\\midrule
Short & 55.67 & 55.56 & 55.67 & 0.23 & 12.54 & 0.59\\
Medium & 43.11 & 42.11 & 42.89 & 0.22 & 12.53 & 0.59\\
Long & 41.56 & 38.78 & 42.44 & 0.23 & 12.54 & 0.6\\
\bottomrule\end{tabular}\end{table*}
\begin{table*}[tp]\centering\small
\caption{Provider frame-budget comparison with the Qwen provider and eight carrier frames. Results are organized by task.}\label{tab:frames}
\begin{tabular}{lrrrr}\toprule
 & \multicolumn{2}{c}{8 provider $\to$ 8 carrier} & \multicolumn{2}{c}{16 provider $\to$ 8 carrier}\\
Task & Accuracy (\%) & Time (s) & Accuracy (\%) & Time (s)\\\midrule
Count & 20.87 & 0.43 & 19.42 & 0.43\\
Ego & 46.88 & 1.90 & 47.44 & 1.99\\
Needle & 49.39 & 0.46 & 48.73 & 0.46\\
Order & 38.22 & 0.71 & 38.22 & 0.67\\
Plot QA & 50.72 & 1.23 & 50.83 & 1.32\\
Anomaly recognition & 62.00 & 0.33 & 63.50 & 0.38\\
Topic reasoning & 79.47 & 1.21 & 80.99 & 0.90\\
Video-MME short & 55.44 & 0.59 & 55.56 & 0.57\\
Video-MME medium & 42.89 & 0.59 & 43.11 & 0.60\\
Video-MME long & 42.22 & 0.60 & 42.67 & 0.60\\
\bottomrule\end{tabular}\end{table*}

\section{Limitations and Discussion}
The reported accuracy changes are small and vary across tasks. The manuscript does not provide sufficient repeated-run statistics to determine their uncertainty. Further evidence is needed for the independent effects of projection, gating, temporal coverage and layer alignment. An increase in provider frame coverage is not by itself evidence that the injection mechanism is superior under equal information or computation budgets.

The text-span interface also depends on implementation-specific decisions: the visual-end marker and separator offset that define the span, tokenizer compatibility between models, and the exchange of post-rotary keys without re-rotation. Models with incompatible tokenizers would require an explicit sequence alignment that is not studied here. In particular, heterogeneous models need not produce semantically aligned representations even when given identical text or similarly sampled frames. Because the injected entries are text-token states, the experiments do not establish which visual information is carried by them; controls that replace the provider's video with mismatched or content-free input are needed to attribute gains to transferred visual content.

Finally, maintaining carrier cache length does not maintain total system memory or computation. The provider and its intermediate states remain additional costs. The available single-provider comparisons do not validate multi-provider scaling, parallel throughput, or generalization to arbitrary model pairs.

\section{Conclusion}
We study communication between multimodal language models through the KV states of shared post-visual text tokens. A provider supplies complementary observations, and projected gated residual updates incorporate them into the corresponding carrier text-token cache entries. The experiments indicate modest, task-dependent accuracy changes and lower measured latency than its text-communication baseline. These findings support further evaluation of KV-state communication of visual context, with explicit alignment rules, matched observation budgets, complete timing boundaries, and uncertainty estimates required for stronger claims.

\end{document}